# Generating Analytic Insights on Human behavior using Image Processing

Namit Juneja, Rajesh Kumar M, *Senior Member, IEEE*
School of Electronics Engineering
VIT University
Vellore, India
namit.juneja2013@vit.ac.in, mrajeshkumar@vit.ac.in

*Abstract*—This paper proposes a method to track human figures in physical spaces and then utilizes this data to generate several data points such as footfall distribution, demographic analysis, heat maps as well as gender distribution. The proposed framework aims to establish this while utilizing minimum computational resources while remaining real time. It is often useful to have information such as what kind of people visit a certain place or what hour of the day experiences maximum activity, Such analysis can be used improve sales, manage huge number of people as well as predict future behaviour. The proposed framework is designed in a way such that it can take input streams from IP cameras and use that to generate relevant data points using open source tools such as OpenCV and raspberryPi. The system uses the AdaBoost Algorithm, HOG detectors and Fisher Faces and just like any other e-commerce platform can generate data points useful in several situations.

## I. INTRODUCTION

Demographic classification has been used by online platforms for a long time, but physical stores often do not enjoy the same benefits as e-commerce platforms because the interaction between the buyer and the seller is more complicated and distributed over different channels. Implementing such a framework on an embedded platform increases its usefulness to several other domains such as human-robot interaction or enhancing face recognition by including gender recognition along with it. Electronic Customer Relation Management (ECRM) is a technology that makes use of such demographic distribution to facilitate product marketing as well as services. The problem of accurately identifying faces in an open space and subsequently their gender has been approached several times in the past. Every method has certain constraints associated with them.

Early methods that were adopted tried to determine the ratio between different features of the face to accurately predict the age of the person [1]. This involved finding all the specific features within the face such as eyes, nose etc, and the finding the distance between them to determine the ratio. This was then used to classify the data into different categories.

There exist other processes that try to depict aging process as a sub space or a manifold. A major constraint in this approach is that the data set required always needs to have full-frontal and aligned faces which is difficult to obtain in the real world scenario. [2]

Most of the implemented methods aim to implement the classification system using powerful machines with ample computational resources. The proposed prototype framework aims to implement a robust classification system that performs in real time at at least 15 frames per second while using limited memory and computational power that the embedded systems offer today. The proposed method also aims to classify data taken from real time input stream such as IP cameras which do not promise an ideal image of the subjects.

## II. DESCRIPTION OF TOOLS

### A. OpenCV

OpenCV which stands for open source computer vision is an image processing framework developed by Intel. Originally written in C it also supports Python, Ruby, C++, Java as well as C#. The primary purpose of the library is its use in real time image processing. We chose this framework as it simplifies the implementation of the algorithms and the respective logic. It works well with the other components of the prototype. We utilize the python wrapper of the library.

### A. Raspberry Pi

Raspberry PI is an open source an open source single board computer that is used for implementing embedded systems based prototypes. We utilize the Raspberry Pi model 3 which comes with a ARM1176JZF-S 700 MHz and a Videocore IV GPU which will be particularly useful in our case. The chipset supports HDMI output and can run linux based operating systems which is necessary in our case. The specifications of the raspberry pi model 3 satisfy our needs from both software as well as hardware perspective.

III. METHODOLOGY

The prototype framework primarily uses three modes to generate different data sets.
Each of them have been described below respectively.

### A. Viola-Jones face detector

The Viola-Jones face detector is used primarily because of 3 main ideas which make it possible to use it for real time face detection. This includes features similar to haar that can be computed by computing an integral image, the AdaBoost Algorithm and Cascade Classifiers which allow classifiers to be combined or quick differentiation between background and foreground parts of the image.[3] [4].

The AdaBoost which is short for Adaptive Boosting Algorithm was first proposed by Schapire and Freund in 1995 [5] which has wide uses in pattern recognition and similar applications. The algorithm is described below.

1. Assume a given sample set $S = (x_1, y_1),........(x_n, y_n)$ $x_i \epsilon X, y_i \epsilon Y = \{-1, +1\}$, number of iterations T
2. We initialize $w_{i,j} = 1 / N$  $i = 1,......,N$
3. Iterating for values of t from 1 ….T
   a. Use $W_t$ distribution to train the weak classifier
   b. The training error for each hypothesis wi is calculated $h_n \mathcal{E}_t = \Sigma W_{t,i} |h_i - y_i|$
   c. Set:
      $a_t = 1/2 \, log(1-\mathcal{E}t / \mathcal{E}t)$
   d. Update the weights:
      $W_{t+1,i} = W_{t,i} exp(-a_t y_i h_t x_i) / Z_t$
      Where $Z_t$ the normalization constant.
4. Hence in the output a stronger classifier is obtained in the final hypothesis
   $H(x) = sign (\Sigma Tt=1 \, a_t h_t(x))$

### B. Gender Determination

The main aim of the algorithm is to identify and hence give most priority to parameters that are useful in classifying the faces into their respective gender. We use supervised learning where the classification algorithm is first trained to a training data containing male and female faces. Then it is used to do further classification. The Eigenfaces method that makes use of Principal Component Analysis is not suitable in this case as it does not give an accurate result. The Fisher faces algorithm gives a specific linear projection which is more suitable for our use case.

1. We generate an image matrix in which every column represents an image. The image matrix is represented by x. A class is assigned to every image with C which is the corresponding class vector.
2. X is projected into dimensional subspace that is N-c and is represented by P. WPca is the rotation matrix that is identified by a Principal Component Analysis. In this
   a. The number of samples in X is N
   b. The number of unique classes is represented with C and is calculated as length(unique(C).
3. The projection P's scatter between the classes is calculated as $\sum_{i=1}^{c} N_i*(m_i - m)*(m_i - m)^T$ which is represented by Sb where
   a. mean of P is m
   b. mean of class i in P is mi
   c. Ni is the number of class samples
4. The Projection P's within class scatter is represented by Sw and is calculated as $\sum_{i=1}^{c} \sum_{kx \in X_i} (kx - m_i) * (kx - m_i)^T$
   where
   a. the samples of class i are Xi
   b. a sample of Xi is kx
   c. mean of class i in P is mi
5. LDA which stands for Linear Discriminant Analysis and the ration of Sw and Sb is maximised. A set of generalized eigenvectors Wfld and their corresponding eigenvalue represents the solution set.
6. The Fisher faces is the product of WFld and WPca.

### C. Heat Maps and Footfall Analysis

1. In order to implement this, we implement a "person detector" using the Histogram Oriented Gradient detector with SVM approach.
2. The Hog detector uses a 64 pixel by 128 pixel image. (Fig 1.)
3. In order the calculate the HOG descriptor we use an 8x8 pixel within the detection window.
4. For each cell, the gradient vector is calculated at each pixel and then represented in a 9 bin histogram.
5. Next, we normalize these histograms by dividing each vector with its respective magnitudes to account for changes in illumination and contrast.
6. The last step is to feed the descriptors created into a support vector machine (SVM) classifier to get an optimum decision function. Once trained the SVM can make decisions if a human body is present in the frame or not. This is later used to identify the regions in a given physical space where the most number of humans visit as well as do footfall analysis by counting the number of humans that pass a given area.

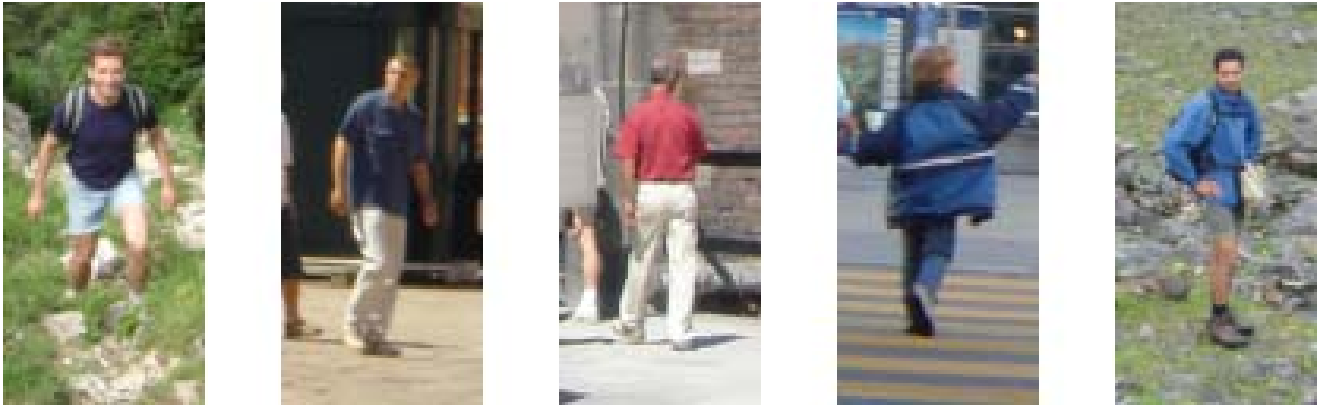

Fig. 1. Examples of detection window from video streams.

IV. IMPLEMENTATION OF THE SETUP

*A. Software Implementation*

OpenCV takes the input video stream from the VGA camera attached to the Raspberry Pi and uses the implemented algorithm to analyse the frame. The code runs in 2 modes.

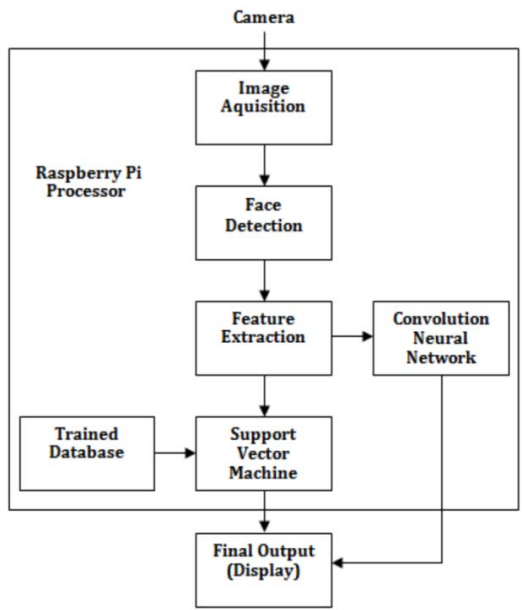

Fig. 2. Block diagram of the proposed system.

1. Gender Determination
   a. If a face is detected in the video the program tries to determine the gender of the image.
   b. Once that is detected it is depicted in the video stream as well as a query is sent to an online firebase server that logs all the data.
   c. A haar cascade classifier is used to detect the face.
   d. The input frame is collected from the input video stream and is stored in a temporary storage.
   e. The frame is passed into a function that tries to detect a face in the image.
   f. If a face is detected it further tries to determine the gender of the face
   g. This keeps running continuously in an infinite loop until the user signals to quit
   h. Wait for the break key

2. Heat Maps and Footfall analysis
   a. The heat maps uses a similar implementation as described above.
   b. A function accepts the frame 15 times every second from the input video stream.
   c. An nxn matrix is maintained with its initial value set to 0.
   d. If a human figure is identified the function updates the cell in the grid where the human figure was located.
   e. Subsequently the colour intensity on the frame for that cell is increased to depict greater human activity in that particular area.
   f. If the coordinates of the human body cross a given user defined line then a counter maintained in the firebase database is updated to keep a count of the number of people present in a given area.
   g. This keeps running continuously in an infinite loop until the user signals to quit
   h. Wait for the break key

3. Graphical User Interface
   a. An online GUI is implemented using HTML and JavaScript.
   b. It uses several graphing libraries to visualize the data stored in the database.
   c. The GUI updates itself in real time as soon as a change is detected in the database.

2017 International Conference on Intelligent Computing and Control (I2C2)

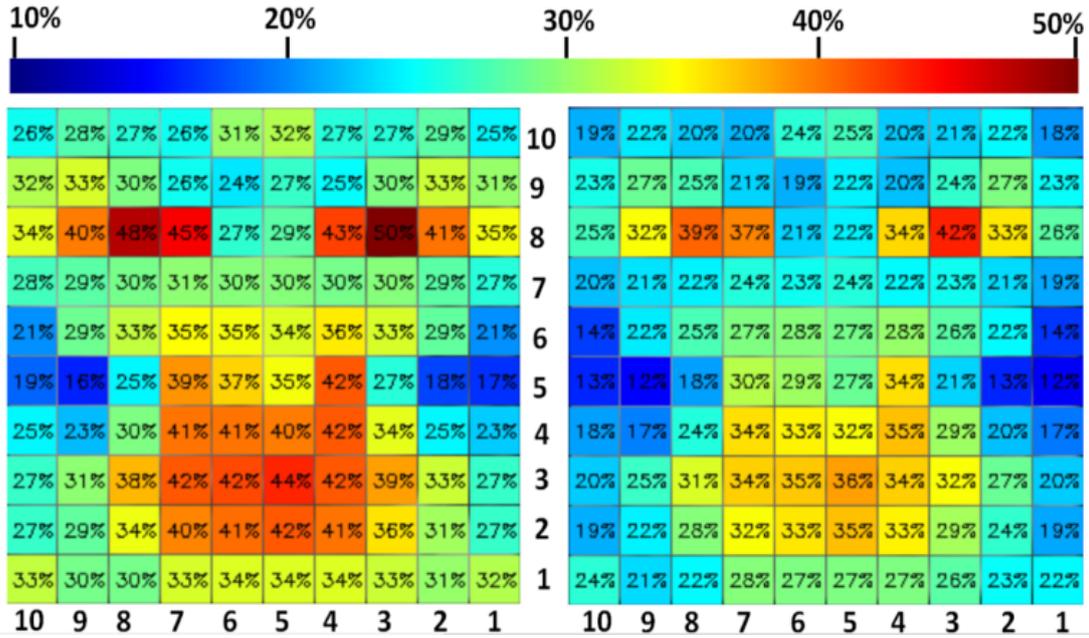

**Fig. 3.** Heat map showing the percentage of people that visited a certain area of the building.

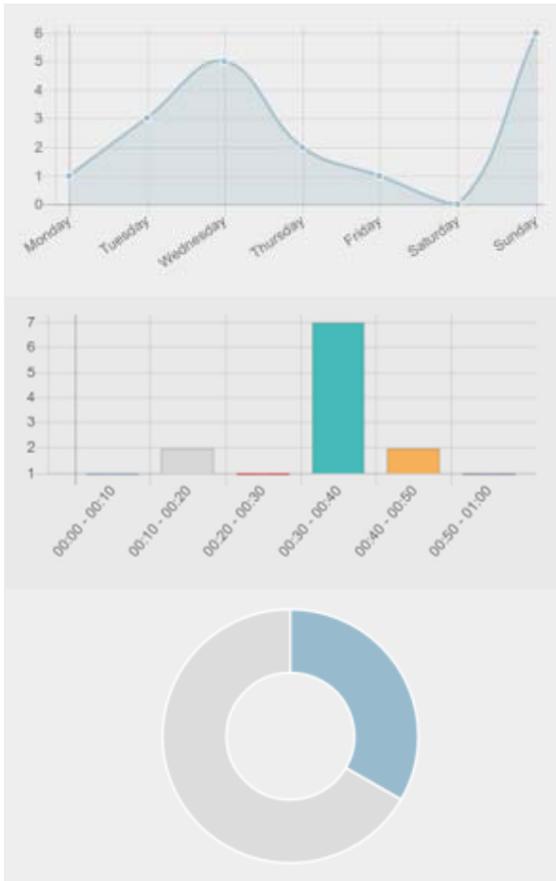

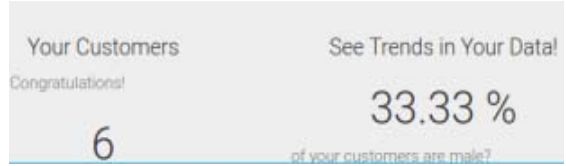

**Fig. 4. Results Dashboard**

V. RESULT AND ANALYSIS

By implementing this prototype framework it was observed that the amount of time required to detect the face is less than 1 second at an average frame rate of 12fps. This was done while keeping the memory usage minimal. From Fig 3. we observe the distribution in terms of the total number of people that visited a given area and how these numbers were distributed over the entire area of the space.

We observe that in Fig 4. that how the frequency of the people entering a physical space varied over different days of the week as well during different hours of the day. The accuracy of such people counting was 82% when the lighting conditions were no appropriate where as in well lit areas the algorithm was able to detect human movements with less than 5% error.

The dashboard also depicts the gender distribution of people entering and leaving a particular physical space and how many people were present at given space during any point of time in the day.

**Table 1.** Gender Detection Result.

| Gender | Accuracy |
|--------|----------|
| Male   | 0.78     |
| Female | 0.91     |

The reason for inaccuracies in the male result is because of several images of children under the age of 10 who have similar facial features to that of female faces. Also, the existence of various faces with makeup, masks, and other distortions increased the classification error. For HOG Detector used for heat maps and footfall analysis, various detection parameters had to be adjusted. The goal was to increase the detection rate by choosing the best parameters while maintaining optimal usage of computational resources and decrease the processing time.

## VI. CONCLUSION

The prototype was implemented and tested successfully. The test results show that this method can be used to effectively determine the demographic distribution of people at a physical space as well as trace human movement while having limited computational complexity. This paper shows that the intersection of image processing and embedded system by using OpenCV and Raspberry Pi is possible while maintaining data output in real time. Future possibilities: Include face recognition along with gender recognition in order to uniquely identify individuals in a physical space.